\documentclass[twoside,leqno,twocolumn]{article}

\usepackage[letterpaper]{geometry}

\usepackage{ltexpprt}

\usepackage{times}
\usepackage{soul}
\usepackage[hyphens]{url}
\usepackage[hidelinks]{hyperref}
\hypersetup{breaklinks=true}
\usepackage{graphicx}
\usepackage{ dsfont }
\usepackage{amsmath}
\usepackage{booktabs}
\usepackage{mathrsfs}
\usepackage{mathtools}
\usepackage{amsmath,amssymb}
\usepackage{subfig}
\usepackage{booktabs,siunitx}
\usepackage[algo2e]{algorithm2e} 
\usepackage[most]{tcolorbox}
\tcbset{
    algotitle/.style={title={\strut
        Algorithm~\thetcbcounter\ifstrempty{#1}{\ignorespaces}{:~#1}}}}
\newtcolorbox[auto counter]{algo}[1][]{%
    colback=white,
    colframe=black,
    boxrule=1pt,
    titlerule=0pt,
    sharp corners,
    left=2pt,right=2pt,top=2pt,bottom=2pt,
    colbacktitle=white,enhanced,
    attach boxed title to top center={yshift=-10pt},
    boxed title style={boxrule=-1pt},
    fonttitle=\bfseries,
    coltitle=black,
    algotitle={},
    #1
}

\DeclareMathOperator*{\Exp}{\mathbb{E}}
\DeclareMathOperator*{\argmax}{arg\,max}

\DeclareMathOperator{\regret}{Regret}
\DeclareMathOperator{\ucb}{UCB}

\begin{document}

\title{\Large An Opportunistic Bandit Approach for User Interface Experimentation \thanks{This work is the result of a collaboration between UC Davis and Target.}}
\author{Nader Bouacida \thanks{Department of Computer Science, University of California, Davis, CA, USA. E-mails: \textbf{\texttt{\{nbouacida, xinliu\}@ucdavis.edu}}}
\and Amit Pande\thanks{Data Science, Target Corporation, Minneapolis, MN, USA. E-mail: \textbf{\texttt{amit.pande@target.com}}}
\and Xin Liu\footnotemark[2]}

\date{}

\maketitle

% Copyright Statement
% When submitting your final paper to a SIAM proceedings, it is requested that you include 
% the appropriate copyright in the footer of the paper.  The copyright added should be 
% consistent with the copyright selected on the copyright form submitted with the paper.
% Please note that "20XX" should be changed to the year of the meeting.

% Default Copyright Statement
\fancyfoot[R]{\scriptsize{Copyright \textcopyright\ 2020 by SIAM\\
Unauthorized reproduction of this article is prohibited}}

% Depending on which copyright you agree to when you sign the copyright form, the copyright 
% can be changed to one of the following after commenting out the default copyright statement
% above.

%\fancyfoot[R]{\scriptsize{Copyright \textcopyright\ 20XX\\
%Copyright for this paper is retained by authors}}

%\fancyfoot[R]{\scriptsize{Copyright \textcopyright\ 20XX\\
%Copyright retained by principal author's organization}}

%\pagenumbering{arabic}
%\setcounter{page}{1}%Leave this line commented out.

\providecommand{\keywords}[1]{\textbf{\textit{Keywords---}} #1}

\begin{abstract} \small\baselineskip=9pt 
Facing growing competition from online rivals, the retail industry is increasingly investing in their online shopping platforms to win the high-stake battle of customer' loyalty. User experience is playing an essential role in this competition, and retailers are continuously experimenting and optimizing their user interface for better user experience. The cost of experimentation is dominated by the opportunity cost of providing a suboptimal service to the customers. Through this paper, we demonstrate the effectiveness of opportunistic bandits to make the experiments as inexpensive as possible using real online retail data. In fact, we model user interface experimentation as an opportunistic bandit problem, in which the cost of exploration varies under a factor extracted from customer features. We achieve significant regret reduction by mitigating costly exploration and providing extra contextual information that helps to guide the testing process. Moreover, we analyze the advantages and challenges of using opportunistic bandits for online retail experimentation.
\end{abstract}


% Introduction ---------------------------------------------------------------------------------------------------------------------------------------------
\section{Introduction}

% The importance of online retail sector 
The global retail sector represents 31\% of the world’s GDP and employs billions of people throughout the globe~\cite{globalRetail}. Much activity in this sector involves traditional retail, but a rapidly-growing fraction comes from online shopping.  Global e-commerce was estimated to be a \$3.53 trillion industry in 2019, marking a gain of 20.7\% from the previous year, according to eMarketer~\cite{ecommerce}. Facing growing competition from online shopping sites, the largest retailers in the world are actively shifting their focus to their online platforms. 

% The  cost of experimenting is dominated by the opportunity cost of providing sub-optimal service to the customers
As with other industries, online retail can be improved by testing new features and through continuous experimentation. Indeed, retailers constantly conduct experiments to improve their online services and platforms~\cite{Brusilovsky2007}. These services are mostly related to search, graphical interfaces, entertainment, retail, and advertising. Online experiments involve changing or creating variations for the service being provided to the clients. Usually, online services can be modified with minimal costs, and any modification of the online user experience can be easily monitored to evaluate its effectiveness. Thus, the cost of experimenting is dominated by the opportunity cost of providing sub-optimal service to the customers.  
%Besides, that blurs the line separating the development process from the production since the online service could be improved after launching it. 

% Problem and motivation
The user interface is an example of online experimentation, which is also one of the most influential factors for attracting customers in the online retail industry. In this paper, we tackle the challenge of reducing the cost of experimenting to identify the most appealing user interface design from a set of designs in online retail platforms. Retail companies are frequently upgrading the user interface of their mobile applications or websites to increase customer engagement and revenue. For many modern platforms, they use randomized trials behind the scenes, known as A/B testing~\cite{abtesting}, to identify alternative user interface designs that result in improvements over the default one. One impediment is the cost incurred during these experiments. During an experiment, customers are shown different variations of a user interface, which may affect overall user experience and reduce customer satisfaction, resulting in lost revenue. Therefore, reducing the cost in such experiments is critical.

%a dramatic increase in the frequency and scope of experiments requires a corresponding reduction in cost.

% Opportunistic Bandits
Multi-Armed Bandits (MAB) are widely used as an improvement to A/B testing~\cite{scott_2015}. MAB make experiments more efficient by steering user traffic towards better-performing variations. In the context of optimizing experimentation for identifying the most popular user interface variation, we study a new paradigm of multi-armed bandits, known as opportunistic bandits~\cite{wu_2018}, where the regret of selecting a suboptimal action varies under a contextual condition called~\emph{load}. The \emph{load} is revealed at each timestep and refers to the price of conducting the trial. When the \emph{load/price} is low, so is the cost/regret of selecting a suboptimal action and vice versa. In our case, we define the \emph{load} as the customer's importance in terms of purchase history or the user's traffic intensity at a particular time interval. As its name suggests, in opportunistic bandits, we leverage the \emph{load} variation as an opportunity to achieve a lower regret. When a customer with a strong purchase history with that retailer arrives, and we should preferably avoid exploration by selecting the arm that we believe is the best.  Such a conservative action is less risky, especially that we already know that taking the risk of selecting a ``bad" action in this situation will result in significant regret. On the other hand, when a customer has low purchase history arrives, trying an uncertain action will result in a lower regret due to the low~\emph{load}. 

% What special about opportunistic bandits?
Opportunistic bandits employ customers' features to guide the exploration-exploitation process, thus achieving better performance. For the sake of demonstrating their superiority, we used three different real datasets provided by Target, one of the biggest retailers in the United States. Our contributions are listed as follows:

\begin{itemize}
  \item We show that opportunistic bandits are naturally aligned with the economics of optimization tests in the online retail industry. 
  \item We evaluate AdaUCB, an opportunistic bandit algorithm, using real datasets and demonstrate that it performs better than other traditional bandits algorithms, achieving up to 78\% regret reduction compared to UCB (Upper Confidence Bound) method.
  \item We analyze the advantages of using bandits algorithms over A/B testing and their limitations and highlight the challenges that we faced in using them. 
\end{itemize}

The remaining of the article is organized as follows: Section~\ref{sec:background} describes the background of A/B testing and multi-armed bandits. Section~\ref{sec:opportunitic} discusses the particular paradigm of opportunistic bandits. Sections~\ref{sec:method} and \ref{sec:evaluation}  discuss the evaluation methodology employed and the experimental results. Section~\ref{sec:discussion} weighs the pros and cons of using opportunistic bandits. Section~\ref{sec:conclusion} concludes.

% Background ---------------------------------------------------------------------------------------------------------------------------------------------
\section{Background} \label{sec:background}
This section provides brief background review about A/B testing and multi-armed bandits.

\subsection{A/B Testing}

A/B testing has been traditionally used to measure conversion rates in static user interfaces variants. The test randomly and evenly splits the traffic between two groups: a control group accessing the existing variation (Version A) and a treatment group accessing the new variation (Version B). We run the test until reaching a sufficient sample size, and then decide whether to implement the winning variation. The latter is decided based on a key metric depending on the website test type: click-through rate (user interface), average time on site (social network), and average checkout time (online shopping), etc. A/B testing can be extended naturally to a scenario with more than two variations, known as A/B/n testing.

A/B tests became very popular among digital marketing companies due to its simplicity. However, researchers~\cite{white_bandit_2012,Hill_2017} outlined several limitations with A/B testing. The fact that an A/B test would allocate an equal amount of traffic to each group means that it not possible to adjust traffic allocation during the test based on what is observed. It ends up wasting resources exploring inferior options in order to obtain statistical significance.

\subsection{Multi-Armed Bandits}
Whereas A/B testing is a frequentist approach, Multi-Armed Bandits (MAB) fall under the bayesian type of methods~\cite{Kuleshov_2014}. It is a hypothetical experiment of a gambler given a slot machine with multiple arms, each with its unknown probability distribution. The objective is to pull the arms in a sequence while gathering information to find the arm with the best expected payout rate and maximize the total payout over the long-run. 

The fundamental challenge is the trade-off between exploiting arms that have performed well in the past to maximize short-run rewards and exploring insufficiently-explored arms, in case they might perform even better~\cite{Auer_2002}. Multi-Armed Bandits testing tries to solve the exploration-exploitation problem using a different approach from A/B testing. Instead of two distinct periods of pure exploration and pure exploitation, bandits interleave both exploration and exploitation in an adaptive fashion.

% Opportunistic Bandits ----------------------------------------------------------------------------------------------------------------------------------------
\section{Opportunistic Bandits} \label{sec:opportunitic}

Our goal is to determine which new user interface design that engage and delight customers with minimal costs. In this context, we may view user interface variations (e.g., test for the best placement of a product display page) in the pool as arms. When a presented variation is clicked, a reward of one is incurred; otherwise, the reward is zero. Therefore, the problem of user interface experimentation can be naturally modeled as a multi-armed opportunistic bandit. Let us consider a test that involves a specific user interface feature with a total of $K$ variants. Following previous work~\cite{wu_2018}, we model it by a $K$-armed stochastic bandit system. Therefore, each arm $k \in 1, \dots, K$ represents a variation receiving user traffic and generates a reward $r_{k,t} \in [0,1]$ at time $t$. The clicks of the customers determine the reward. When a variation is presented to a customer, if he clicks, a payout of 1 is incurred and 0 otherwise. The true CTR $c_k = \Exp(r_{k,t})$ for the arm $k$ is the mean value of the rewards~\footnote{The CTR of an arm is not associated with a user and aggregated across all traffic.} and is used to determine the best user interface feature variation. Let $c^* = \max_k c_k$ be the maximum expected value (maximum true CTR). The goal of the experiment is to find the feature variation (the arm) with the highest CTR, denoted by arm $k^* = \argmax c_k$.

In an ideal world, we would already know all arms' probability distribution, and thus apply all our resources towards that one action that generates the greatest return. Unfortunately, that is not the world we live in, and the arms' values are considered unknown. Hence, we need to estimate the CTR values of all arms, and as such, need to maximize our ability of that discovery. Bandit algorithms try to minimize exploration costs by moving traffic towards winning variations gradually, instead of waiting for a ``final answer" at the end of an experiment. 

 With this definition of the reward, the expected payoff of an interface variation is precisely its Click-Through Rate (CTR), and choosing a variation with a maximum CTR is equivalent to maximizing the expected number of clicks from customers. Therefore, opportunistic bandits add another dimension to the traditional multi-armed bandits by using a factor measuring customer importance (this will be referred to as \emph{load} $L_t \geq 0$ at time $t$) to guide the exploration-exploitation process. Considering our problem, the customer prior 90 days purchased items total value is an appropriate choice for \emph{load} $L_t$ because it captures the customer importance from a financial point of view. In fact, we select customer importance as \emph{load} because we aim to minimize lost revenue resulting from allocating highly-important customers to poorly-performing variations of the test. In our bandit formulation, we maximize the sum of the rewards modulated by the \emph{load} factor.
 
 %, which in turn is the same as maximizing the total expected reward in our bandit formulation. 

AdaUCB, an opportunistic bandit algorithm, is implementing this paradigm by defining a \emph{load}-dependent exploration term when calculating the upper-confidence bound for each arm in Equation~(\ref{eq:ucb}). The agent observes the value of the \emph{load} $L_t$ at time $t$ before making the decision. Based on normalized \emph{load} $\tilde{L}_t$ and historical observations in previous trials $\mathcal{H}_{t-1}$, our agent policy $\pi$ pulls an arm $a_t = \pi(\mathcal{H}_{t-1}, \tilde{L}_t)$. The agent then receives an actual reward $L_t \, r_{a_t,t}$. Unlike nominal reward $r_{a_t,t}$, the actual reward is dependent on the \emph{load} $L_t$. This model sheds lights on the general concept of opportunistic bandits. The nominal reward $r_{a_t,t}$ will capture the clicks. Meanwhile, the actual reward, the click monetary reward, is modulated by the real \emph{load} $L_t$ as $L_t \, r_{a_t,t}$ (customer purshasing power $\times$ click reward). In an ideal scenario, the agent have prior knowledge about all arms' probabilities distributions and will always pull the best arm obtaining the total reward $c^* \Exp \left( \sum_{t=1}^{T} L_t \right)$. Hence, the regret of policy $\pi$ can be expressed as:

\begin{equation} \label{eq:regret}
    \regret_{\pi}(T) = c^* \Exp \left( \sum_{t=1}^{T} L_t \right) - \sum_{t=1}^{T} \Exp \left( L_t \, r_{a_t,t} \right) 
\end{equation}
We note that we use normalized \emph{load} $\tilde{L}_t$ defined in Equation~(\ref{eq:normalization}) instead of real \emph{load} $L_t$ in calculating Upper Confidence Bounds values in Equation~(\ref{eq:ucb}) in order to capture the real \emph{load} different ranges:

\begin{equation} \label{eq:normalization}
    \tilde{L}_t = \frac{[L_t]_{l_{\min}}^{l_{\max}} - l_{\min}}{l_{\max} -l_{\min}}
\end{equation}
where $l_{\min}$ and $l_{\max}$ are respectively the lower and the upper thresholds for truncating the \emph{load} level, and $[L_t]_{l_{\min}}^{l_{\max}} = \max\{l_{\min}, \min\{L_t,  l_{\max}\} \}$. 

Formally, AdaUCB algorithm proceeds in discrete trials $t = 1, 2, \dots$. In trial $t$ (the trial scope is one user session) :

\begin{enumerate}
  \item The algorithm observes the current user $u_t$ and his corresponding load $L_t$. 
  \item Based on observed payoffs in previous trials, it calculates the Upper Confidence Bound (UCB) for each arm $k \in 1, \dots, K$ as follows: \\
  \begin{equation} \label{eq:ucb}
      \ucb_{k}(t) = \hat{\mu}_k(t) + \sqrt{\frac{\alpha \, (1- \tilde{L}_t) \, log(t)}{n_{k,t-1}}}
  \end{equation}
  where $\hat{\mu}_k(t) = \frac{1}{n_{k,t-1}} \sum_{t=1}^{T} \mathds{1} (a_t = k) \, r_{k,t}$ is the posterior mean (the exploitation factor), $n_{k,t-1}$ is the number of click counts of the arm $k$ prior to time $t$ and $\alpha$ is a parameter of the algorithm controlling the confidence interval width. The convergence guarantees of AdaUCB algorithm have been proven in~\cite{wu_2018}.
  
  \item The algorithm chooses the arm $a_t$ with the largest UCB value:
  \begin{equation*}
      a_t = \argmax_{1 \leq k \leq K} \left ( \ucb_k(t) \right )
  \end{equation*}
  \item The algorithm receives an actual payoff $L_t r_{a_t, t}$ , whose expectation depends on both the user $u_t$ and the arm $a_t$. It improves its arm-selection strategy with the new observation by updating $\hat{u}_k(t)$ and $n_{k,t}$.
\end{enumerate}

The AdaUCB algorithm adjusts the trade-off between exploration and exploitation based on the normalized \emph{load} level $\tilde{L}_t$. It makes decisions based on the posterior mean (the exploitation term) $\hat{\mu}_k(t)$ and the confidence interval width (the exploration term). AdaUCB employs an exploration factor that is linearly decreasing in terms of \emph{load}. As a result, when the \emph{load} is low, so is the cost/regret of picking a suboptimal action and vice versa. Intuitively, we should explore more when the \emph{load} is low and exploit more when the \emph{load} is high. As its name suggests, in opportunistic bandits, we leverage the opportunities of \emph{load} variation to achieve a lower regret. This paradigm is naturally aligned with user interface experimentation, as discussed earlier.

Generally speaking, opportunistic bandits can be viewed as a particular case of contextual bandits~\cite{Tyler2010} where the context is provided by \emph{load}. However, applying existing general contextual bandits algorithms will not generate optimal regrets because they do not take advantage of the unique properties of opportunistic bandits; in particular, the optimal arm remains the same throughout the experiment, and regrets differ under different contexts (i.e., \emph{load}) as shown in~\cite{wu_2018}. The exploration cost varies under different environmental conditions, and not the best arm given the context. 

% Evaluation Methodology ----------------------------------------------------------------------------------------------------------------------------------------
\section{Evaluation Methodology} \label{sec:method}

Usually, the experiments conducted by retail companies are performed in an online fashion. Instead, the data that we have is an offline data that was collected at a previous time using an entirely different logging policy (A/B testing). Rewards are only observed for the actions chosen by the logging policy, which are likely to often differ from those chosen by the algorithm being evaluated. Therefore, it is not clear how to proceed with evaluation based on offline data~\cite{li_2011}.  

Building a simulator to mimic the bandit process from the logged offline data is one possible solution to consider. However, simulator-based evaluation is considered an unreliable solution because it introduces a bias in the evaluation approach. For this purpose, we use the offline evaluation technique proposed in~\cite{li_2010}. It is demonstrated to be an unbiased method given two assumptions are met: the individual events of the data log are independent and identically distributed, and the logging policy used to collect the logged data selects each arm at each time step uniformly at random. We note that the second assumption can be considerably weakened so that any randomized logging policy is allowed, but at the cost of decreased efficiency in using the data log. 

\begin{algo}[algotitle=Evaluation Algorithm, label=evaluation]
\SetAlgoLined
\vspace{5pt}
Input: $T>0$; policy $\pi$; events log $\mathcal{E}$ \\
Initialize $i = 0$ \\
%\KwData{Confidence set constructor $\Phi$}
\For{$t=1,2,\dots, T$}{
    \Repeat{$a_t = a$}{
      Get next event $(L, \tilde{L}, a, r_a)$\\
      $a_t = \pi(\mathcal{H}_{t-1}, \tilde{L})$\;
    }
    Update $\mathcal{H}_{t} = \mathcal{H}_{t-1} \cup (L, \tilde{L}, a, r_a)$ \\
    Update $R_t = R_{t-1} + r_a$ 
}
\end{algo} 

We posit access to a large sequence of logged events $(L, \tilde{L}, a, r_a)$ generated by the interaction of the logging policy (A/B testing) with the environment. Each such event consists of the real \emph{load} $L$, the normalized \emph{load} $\tilde{L}$, a selected arm $a$ and the observed reward $r_a$. Crucially, only the reward $r_a$ is observed for the single-arm that was chosen uniformly at random. We aim to use this data to evaluate a bandit algorithm $\pi$. Formally, $\pi$ is a (possibly randomized) mapping for selecting the arm at a time $t$ based on the history $\mathcal{H}_{t-1}$ of $t-1$ preceding events, together with the current normalized \emph{load} $\tilde{L}$.

The evaluation method discussed above is described in Algorithm~\ref{evaluation}. It takes as input the desired number of ``processed" or ``valid" events $T$, a bandit algorithm $\pi$ and the data log on which to base the evaluation. It proceeds by stepping through the stream of logged events one by one. If given the current history $\mathcal{H}_{t-1}$ and the normalized \emph{load} $\tilde{L}$, our algorithm policy $\pi$ selects the same action as the one that was selected by the logging policy, then the event is retained, and the total reward $R_t$. Otherwise, if the algorithm in question chooses a different action from the one that was taken by the logging policy, then the current event is ignored, and we proceed to the next event in the data log without any change. For the sake of simplicity, we assume that the log stream is infinite. In practice, we simply cycle through the data log if we reach its end. Therefore, with a large enough log and a reasonable number of desired events $T$, the evaluation algorithm is unlikely to encounter this problem. Because the logging policy chooses each arm uniformly at random, each event in the log is retained by this algorithm with a probability exactly $1/K$ where $K$ is the number of arms, independently of anything else. Therefore, the events which are retained have the same distribution as if they were generated by the unknown distribution $D$, from which tuples consisting of observed \emph{load} and hidden payoffs for all arms are drawn independent and identically distributed.

%In the subsequent experiments, We draw the attention of the reader that the competing bandit algorithms make the following mathematical assumptions:

%\begin{itemize}
%  \item Conversion rates\,\footnote{True Click-Through Rate of each arm.} does not change over time.
%  \item Displaying a variation and observing a conversion happen instantaneously. 
%  \item Samples in the bandit algorithms are independent of each other.

%\end{itemize}

\section{Experiments} \label{sec:evaluation}

This section gives a detailed description of our experimental setup, including data pre-processing, performance evaluation, and competing algorithms.

\subsection{Data Pre-processing}
The original dataset is a raw data that contains clicks and displays captured during A/B testing process for three different tests:

\begin{itemize}
  \item First Test: We are testing for the best placement for a product display page. There are three possible placements (3 arms). The raw data contains 17 days of collected events in August 2018. After pre-processing, we keep 3.4 million valid events. 
  \item Second Test: We are testing to identify the best strategy for the retailer's mobile application home page. Similar to the first test, we are experimenting with three variants (3 arms). The raw data contains collected events in July and August 2018. After pre-processing, the final log has 1.1 million valid events.
  \item Third Test: We are trying to determine whether a new skin of search filters is better than the old ones. Hence, this test contains two variants (2 arms). The raw data contains 19 days of collected events in June 2018. After pre-processing, the final log contains 3.7 million events.
\end{itemize}

The dataset contains several features such as server timestamp, customer identifier\,\footnote{The data was anonymized to protect customer private or sensitive data.}, customer purchasing power, action clicked or displayed, and session hit. The customer purchasing power represents his ability to make purchases and is estimated based on the total value in dollars of the items purchased in the last 90 days. The session hit is the sequential number of the visited pages within a session. We need to pre-process the data to remove duplicates and discard invalid entries. The first step was to clean the clicks and displays in the data frame and remove duplicates. Since the server timestamp is not accurate enough, it is not clear how to correctly join the clicks to their corresponding displays. We recall that the data provides us with the session hit. Therefore, we join clicks and displays based on customer identifiers, arm displayed or clicked, the date, and session hit.  Then, we filter clicked entries using session hit based on the condition ``\emph{session hit of display $\leq$ session hit of click}". Indeed, a click event cannot take place before a matching display. After that, we include the rewards based on whether the entries are just displayed and not clicked (we assign a reward of zero) or displayed and clicked (we assign a reward of one).  Based on the customer identifier, we calculate the normalized \emph{load} and include it in the processed data along with the real \emph{load} (customer purchasing power). Customers with no purchase history are assigned the average \emph{load} across all users. Unless otherwise stated, we define the truncation thresholds for \emph{load} normalization such that $\mathds{P}(L_k \leq l_{min}) = \rho$ and $\mathds{P}(L_k \geq l_{max}) = \rho$ where $\rho = 0.05$.

\subsection{Experimental Results}
With the introduction of the cloud and tools such as Google Analytics~\cite{tayler2006google}, online retailers can conduct experiments quicker and easier than ever before. Indeed, they can experiment continuously, perpetually improving different aspects of their offerings. Because we are only limited to offline data, we use the methodology described in section~\ref{sec:method} for evaluation purposes. The algorithms empirically evaluated in our experiments are the following:

\begin{itemize}
  \item Decaying Epsilon Greedy (DEG): It estimates each arm's CTR; then, it chooses the arm with the highest CTR estimate (the action that seems best at that moment) with probability $1-\epsilon$, and it selects a random action with probability $\epsilon$. It tries to decrease the percentage dedicated for exploration $\epsilon$ as time goes using annealing. This algorithm can give near-optimal regret as \emph{Auer et al.} demonstrated that a tuned $\epsilon$-greedy will almost always outperform their UCB algorithm (section 4.1 of~\cite{Auer_2002_1}). We calibrate the discount factor of $\epsilon$.
  \item Thompson Sampling (TS): An algorithm that draws a sample CTR from each arm's Beta distribution, and pulls the arm with the highest sampled CTR. 
  \item UCB: This algorithm estimates each article's CTR as well as a confidence interval of the estimate, and always chooses the arm with the highest UCB following UCB1~\cite{Auer_2002}. The only parameter of this policy is $\alpha$, which is calibrated empirically.
  \item AdaUCB: Opportunistic algorithm described in section~\ref{sec:opportunitic}. The only parameter of this policy is $\alpha$, which is calibrated empirically. We note that AdaUCB is the opportunistic variant of UCB1.
  \item Optimal: A policy that always selects the best arm. This represents the ideal case where the algorithm knows the true CTR of each arm. 
\end{itemize}

\begin{figure*}[t] 
	\centering
	\subfloat[\emph{First Test} \label{rs1}]{\includegraphics[width=0.333\textwidth]{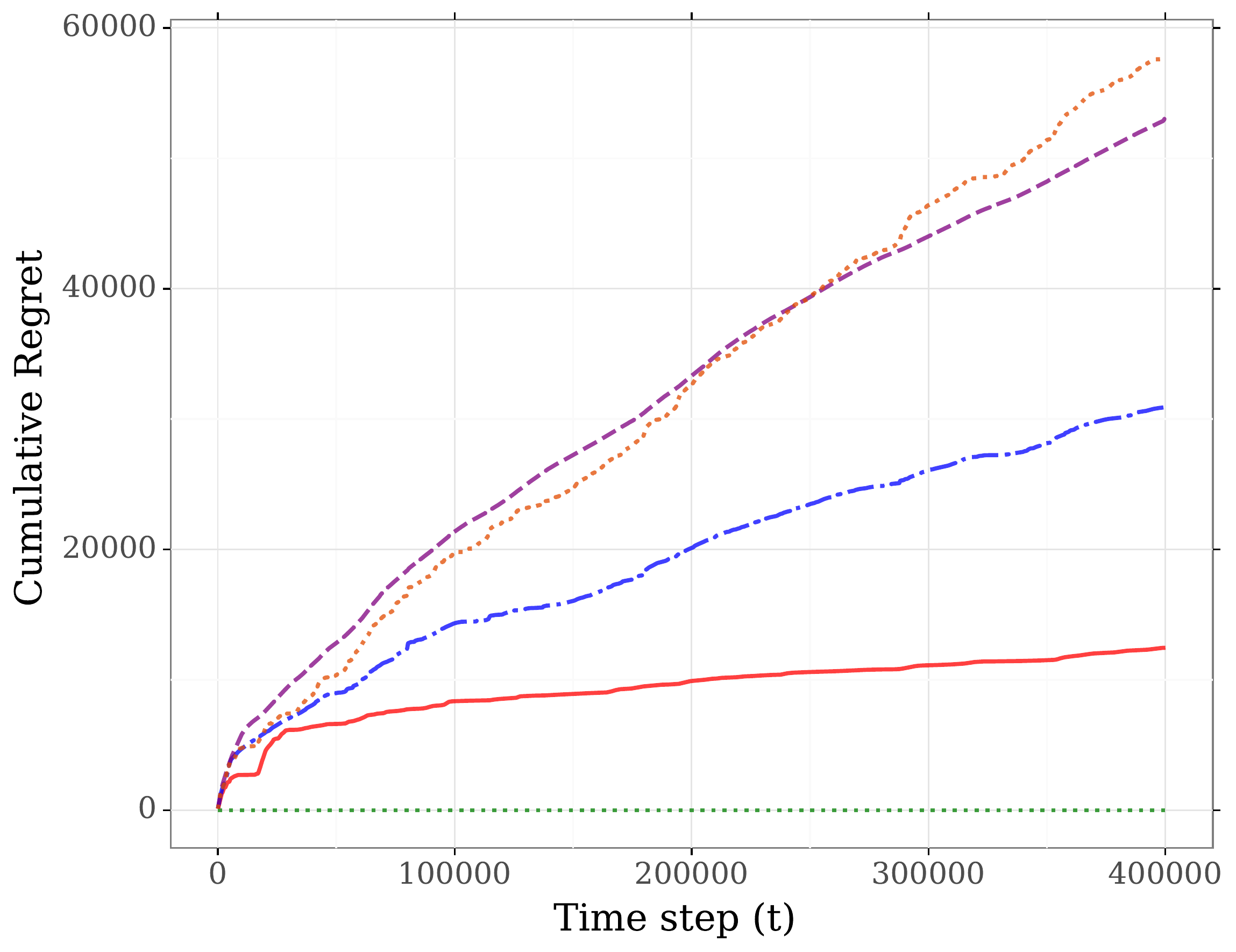}}\hfill
	\subfloat[\emph{Second Test} \label{cw1}]{\includegraphics[width=0.334\textwidth]{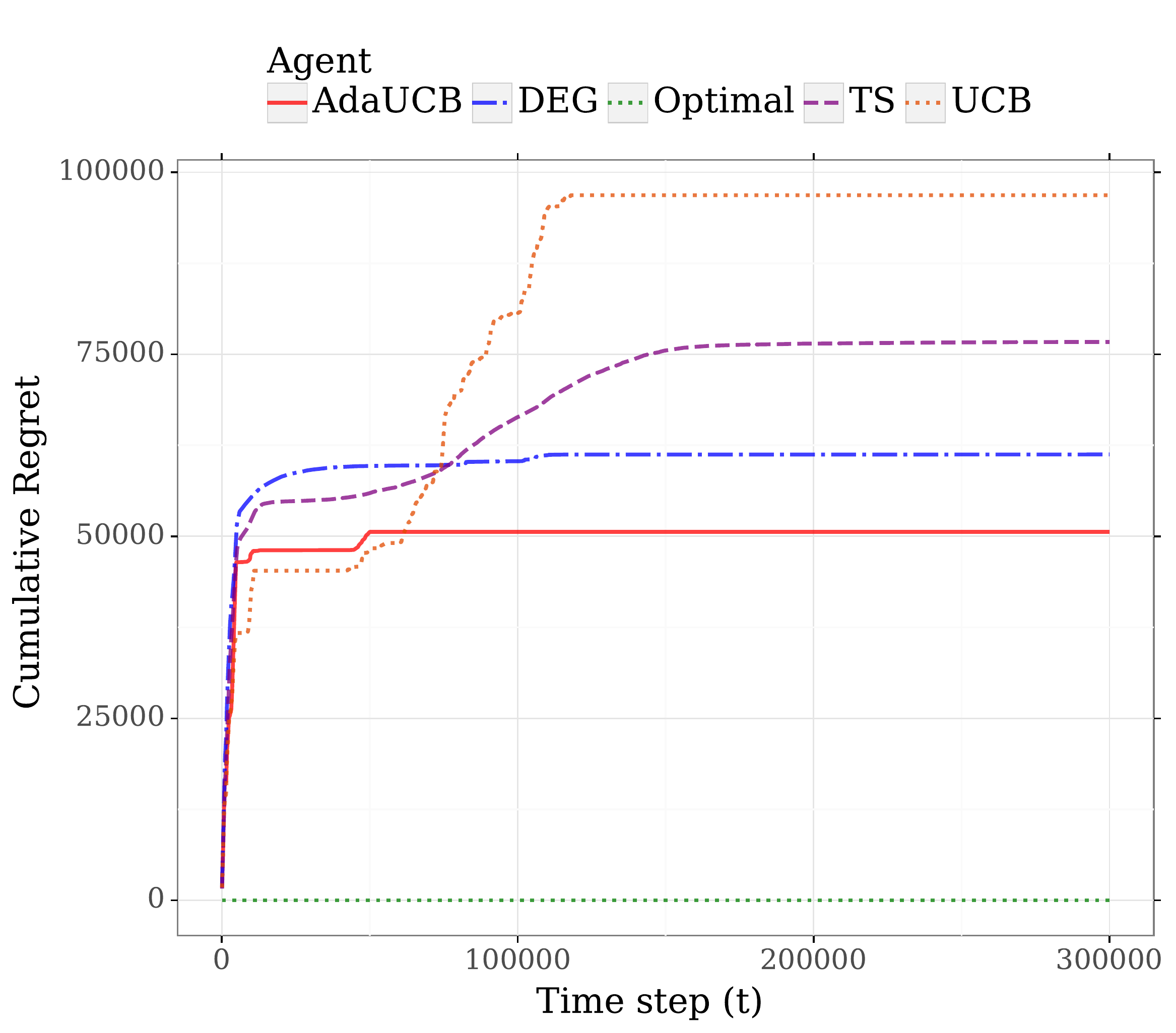}}\hfill
	\subfloat[\emph{Third Test} \label{fl1}]{\includegraphics[width=0.333\textwidth]{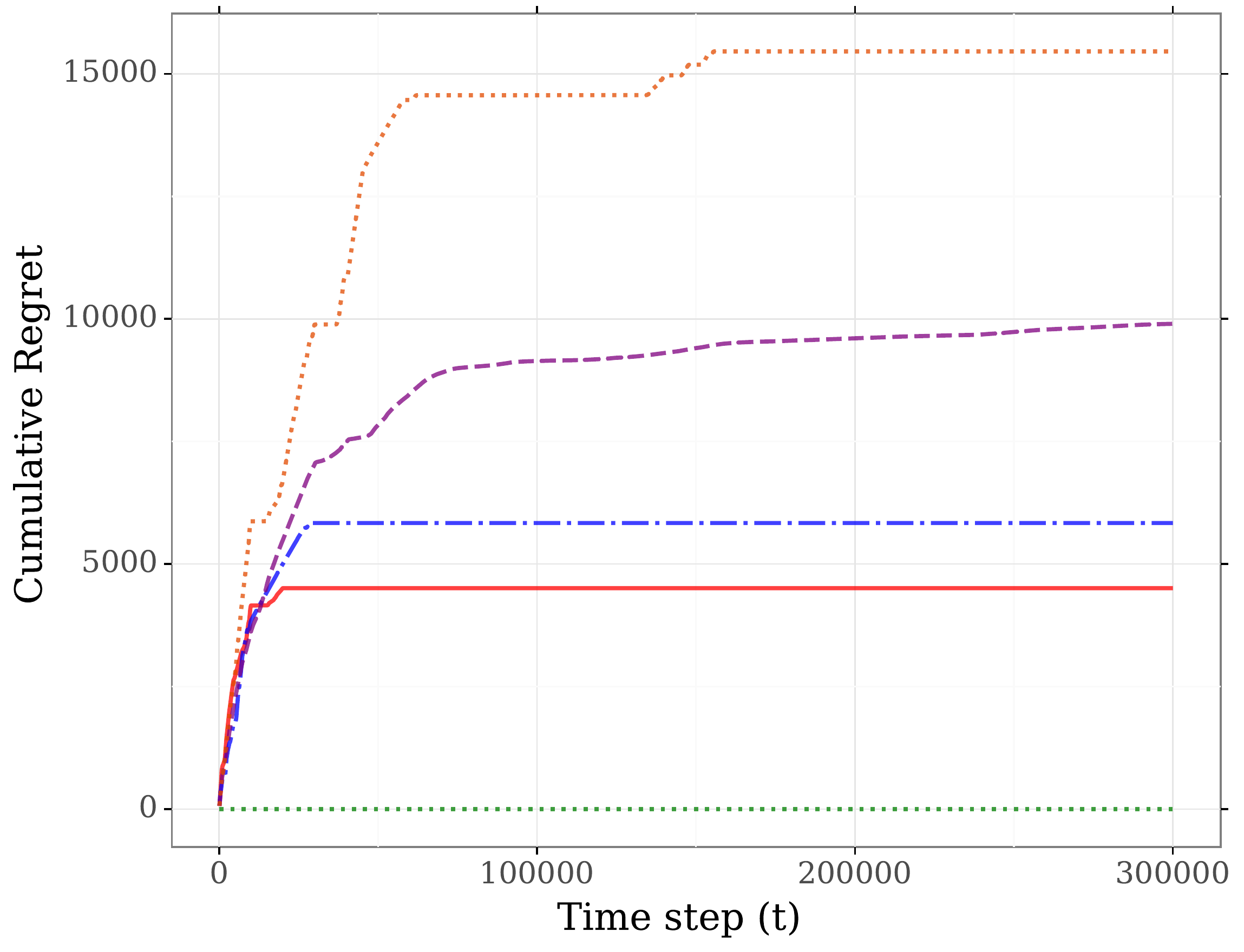}}\hfill
	\caption{Cumulative Regret (First Scenario)}
	\label{fig1}
\end{figure*}

\begin{figure*}[!h]
	\centering
	\subfloat[\emph{First Test} \label{rs2}]{\includegraphics[width=0.333\textwidth]{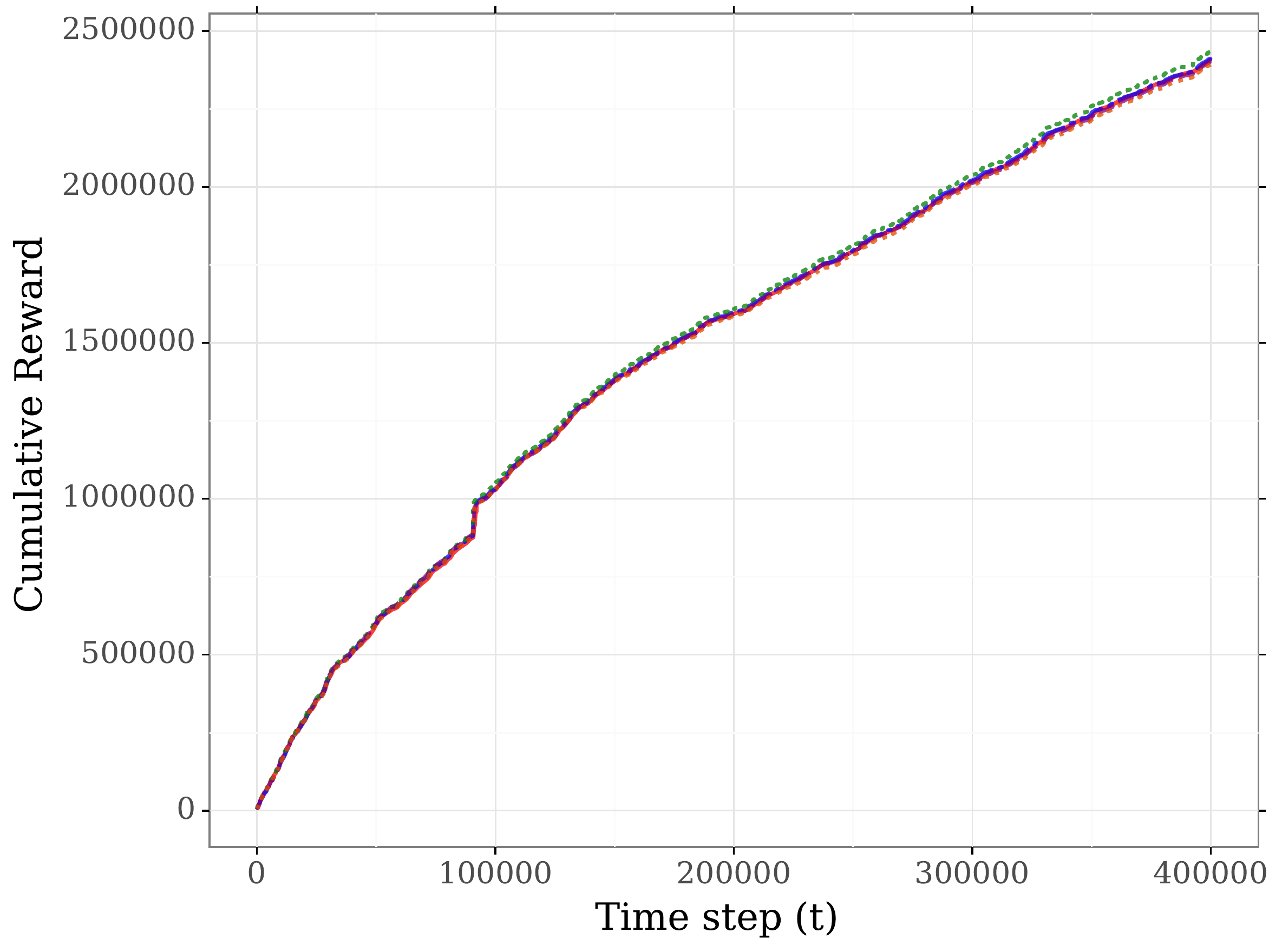}}\hfill
	\subfloat[\emph{Second Test} \label{cw2}]{\includegraphics[width=0.334\textwidth]{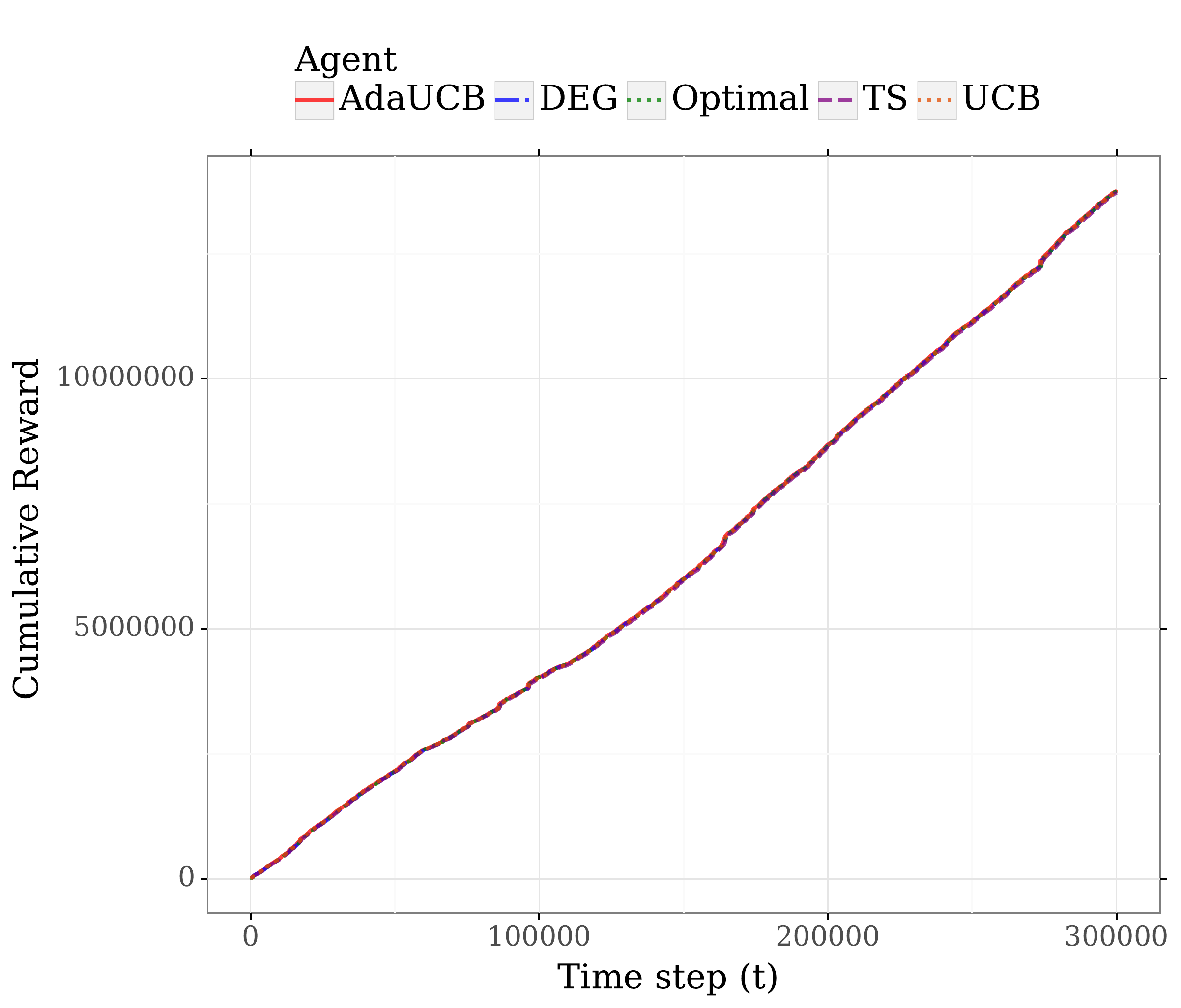}}\hfill
	\subfloat[\emph{Third Test} \label{fl2}]{\includegraphics[width=0.333\textwidth]{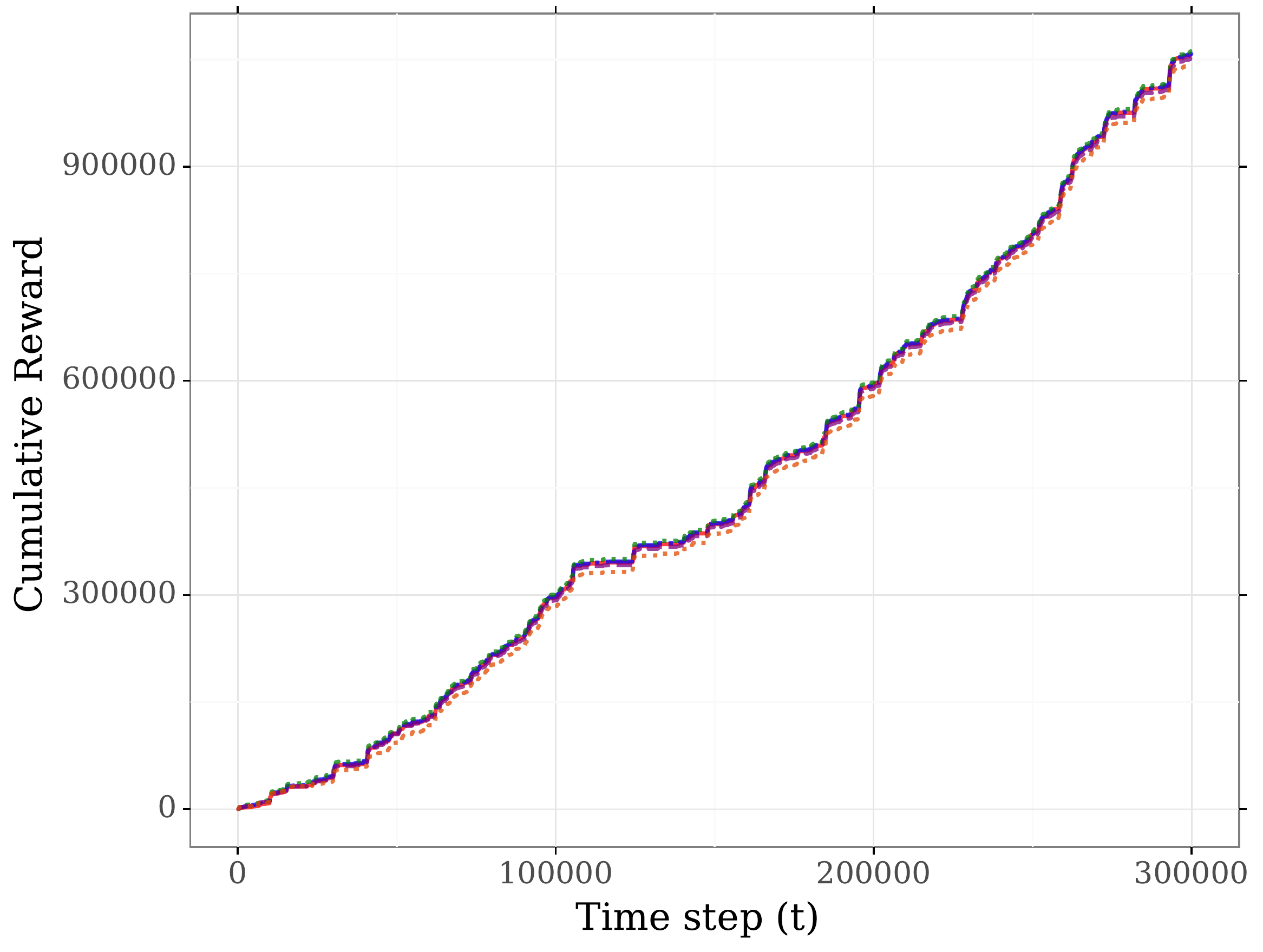}}\hfill
	\caption{Cumulative Number of Clicks (First Scenario)}
	\label{fig2}
\end{figure*}

To evaluate these algorithms, we run each algorithm using 1000 seeds and compute the average cumulative regret and the average cumulative reward. As illustrated in Figure~\ref{fig1} and Table~\ref{tab:first}, AdaUCB achieves lower regret than its competitors across all three tests, outperforming DEG, UCB, and Thompson Sampling (TS). For instance, AdaUCB attains respectively 59.69\%, 17.36\%, and 22.77\% lower regret than Decaying Epsilon Greedy (DEG) for the three tests. As expected, when perfectly tuned, DEG is performing better than UCB and TS. UCB suffers from considerable regret. AdaUCB, the opportunistic variant of UCB, successfully reduced the regret of UCB, respectively, by 78.41\%, 47.74\%, and 70.83\% for the three tests. Opportunistic bandits allocate fewer resources to inferior arms when the \emph{load} is high.  When the \emph{load} is low enough, AdaUCB encourages the exploration of inferior under-explored arms. The lower customer purchasing power minimizes the cost of exploration. Alternatively, AdaUCB assigns the fewer users with solid buying power to track the temporal changes of the best-performing variations. Hence, traffic is selectively allocated towards the best-performing or poorly-performing arms, reducing the overall regret. Opportunistic bandits create a natural way to automate the users' traffic allocation. As the experiment progress, AdaUCB learns more and more about the relative payoffs, and by doing so, it does a better job in selecting useful variations. The gain resulting from introducing the \emph{load} factor is substantial, showing the potential of opportunistic bandits. 

\begin{table}[h]
\setlength{\tabcolsep}{5pt}
\centering
\begin{tabular}{@{} l ccc @{}} % @{} serves to suppress white space at ends of table
\toprule
Algorithm   &  \multicolumn{3}{c @{}}{Datasets}\\ 
\cmidrule(l){2-4}
    & {First Test} & {Second Test} & {Third Test}  \\
\midrule
A/B Testing & 733760 & 3546618 & 165375 \\
DEG     & 30922 & 61242 & 5836  \\
TS      &  53048 & 76697 & 9899 \\
UCB     &  57744 & 96847 & 15456 \\
AdaUCB  &  12464 & 50610 & 4507  \\
\bottomrule
\end{tabular}
\caption{Accumulated Regret at the end of the experiments (First Scenario).}
\label{tab:first}
\end{table}

A/B testing (not shown in the figures) has linear regret, while the bandit algorithms achieve logarithmic regret. Table~\ref{tab:first} shows that bandit algorithms outperform A/B testing by at least one order of magnitude. Generally speaking, the multi-armed bandit approaches are dramatically more efficient at finding the best arm than traditional statistical experiments because they move traffic towards winning arms gradually, instead of waiting for a ``final answer'' at the end of an experiment. They converge faster because samples that would have gone to obviously inferior arms can be assigned to potential winners. The extra data collected on the high-performing variations can help distinguish the ``good'' arms from the ``best'' ones more quickly. 

\begin{figure*}[t] 
	\centering
	\subfloat[\emph{First Test} \label{rs3}]{\includegraphics[width=0.333\textwidth]{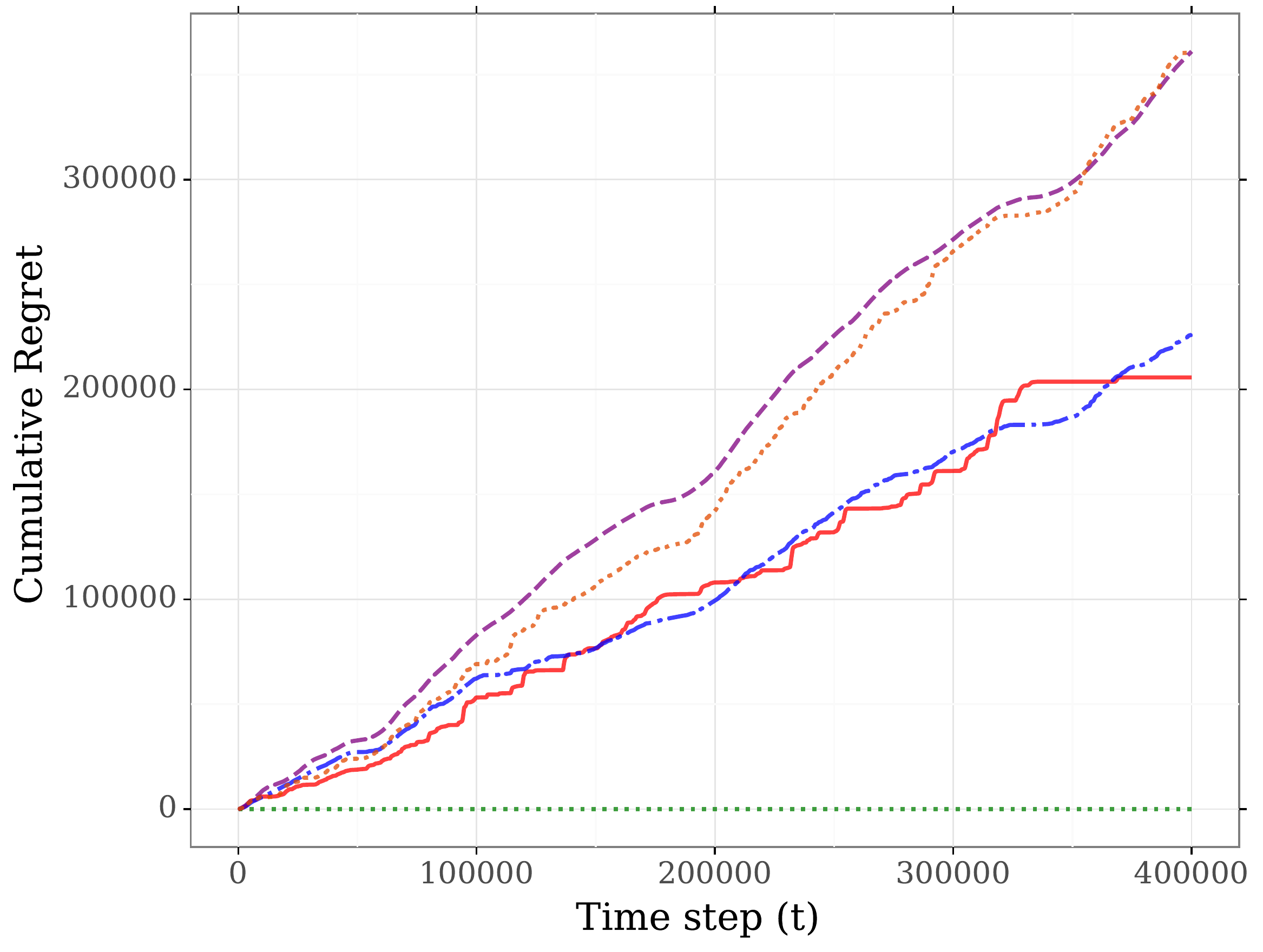}}\hfill
	\subfloat[\emph{Second Test} \label{cw3}]{\includegraphics[width=0.334\textwidth]{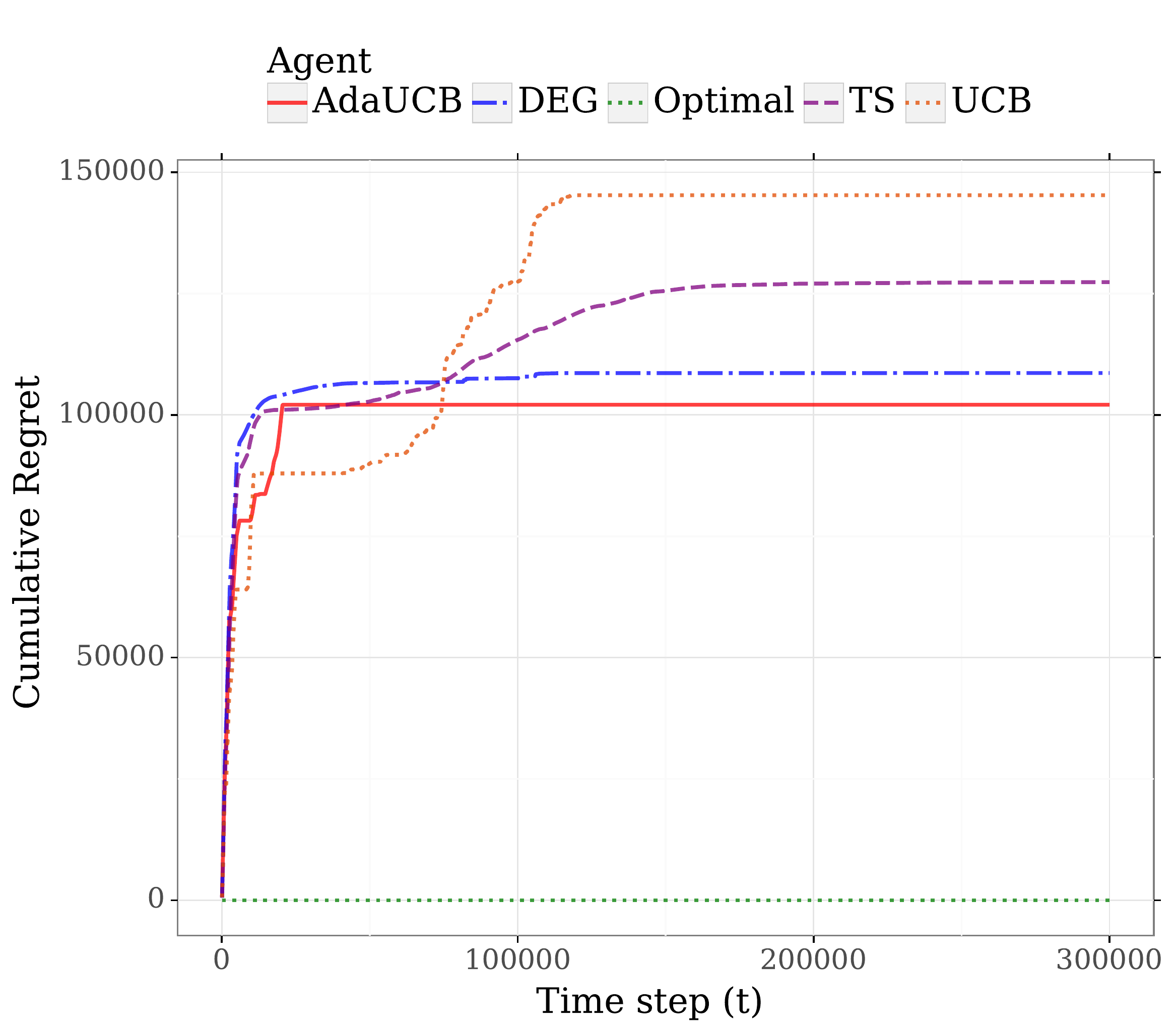}}\hfill
	\subfloat[\emph{Third Test} \label{fl3}]{\includegraphics[width=0.333\textwidth]{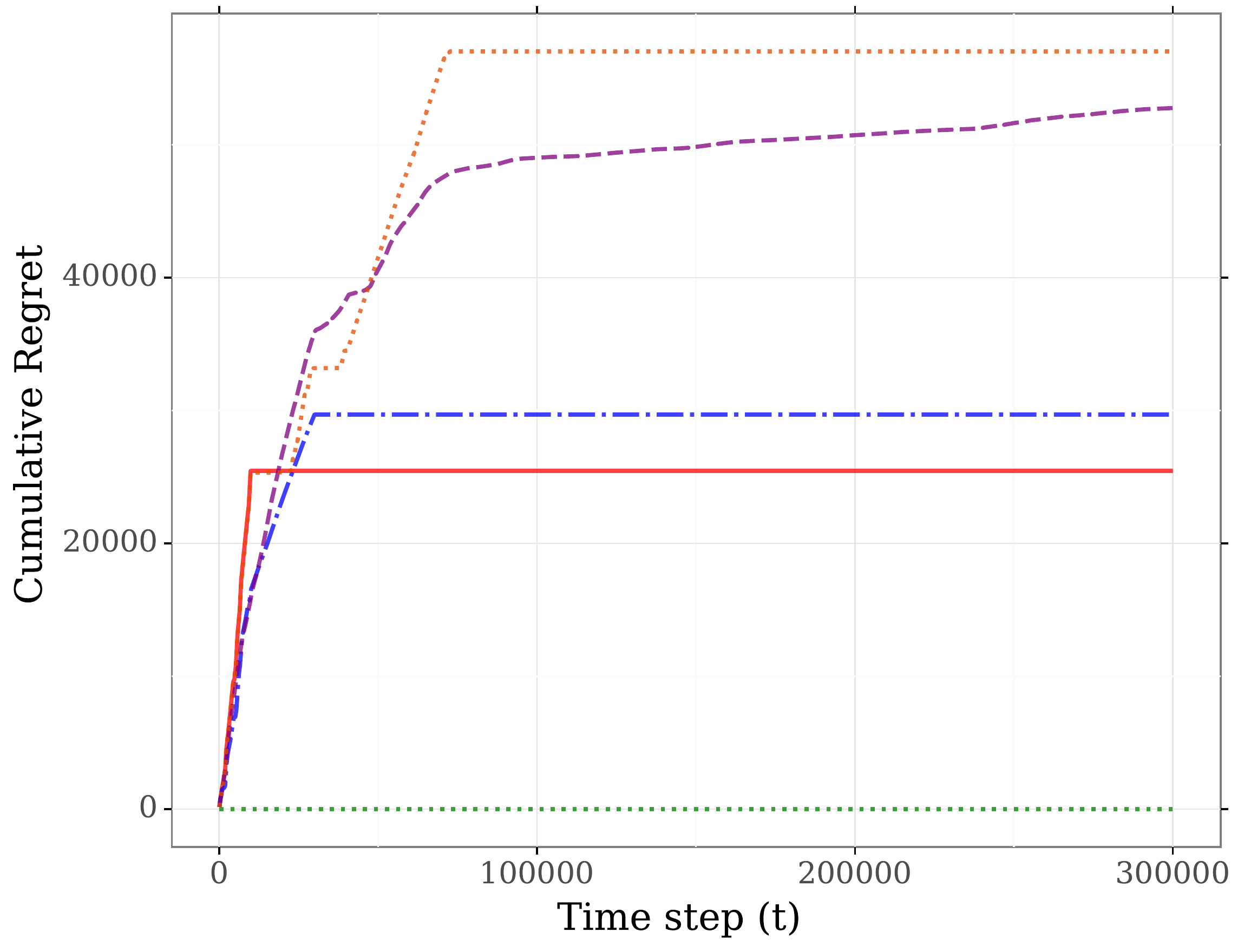}}\hfill
	\caption{Cumulative Regret (Second Scenario)}
	\label{fig3}
\end{figure*}

\begin{figure}[!h] 
	\centering
	\includegraphics[width=\columnwidth]{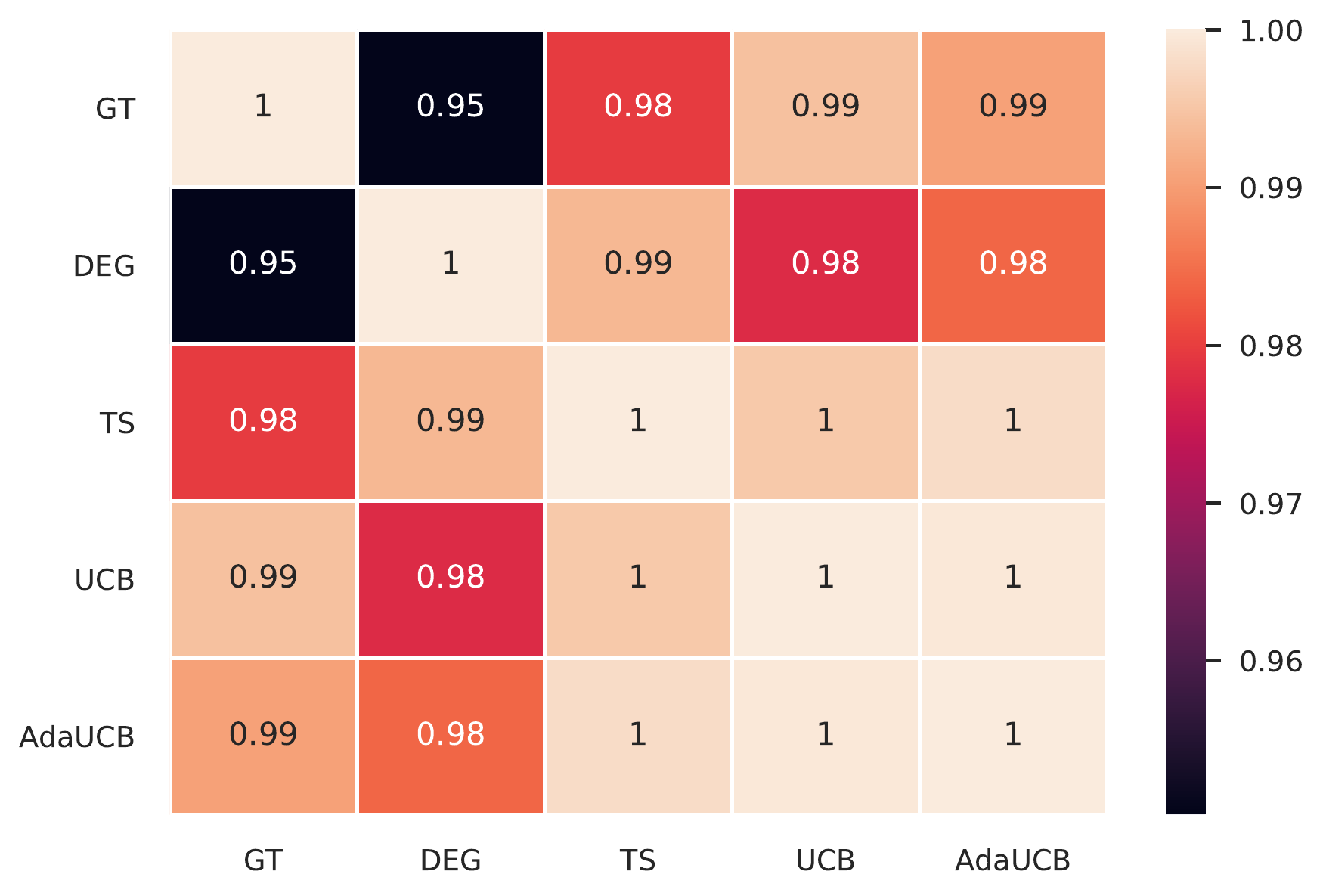}
	\caption{Correlation Matrix for the arms' CTRs values when compared to the Ground Truth (GT) (First Scenario)}
	\label{fig4}
\end{figure}

To better understand the effectiveness of our approach, we visualize the correlation matrix of the estimated CTRs for each method and the true CTRs representing the Ground Truth (GT). Figure~\ref{fig4} shows the existence of a strong correlation between the Ground Truth (GT) and all algorithms estimated CTRs. UCB and AdaUCB achieve the best correlation with the Ground Truth (GT), suggesting that they generate the most accurate CTRs estimates. This comes at the cost of higher regret for UCB, resulting from the increased exploration compared to the other policies. AdaUCB offsets this cost by avoiding expensive exploration. The correlation scores across different algorithms appear very close because we are experimenting with data that have a small number of arms with very low CTRs. Even a minor improvement will be considered significant in such conditions.

Moreover, we investigate the cumulative number of clicks collected by each algorithm over time. As shown in Figure~\ref{fig2}, We notice that all methods collect a similar cumulative reward over time, through all three tests. However, we expected to produce a larger cumulative reward for AdaUCB, given its proven superiority in terms of regret minimization. In order to examine this observation, we evaluate another method, the Optimal policy, which always selects the best arm. Practically, we cannot reach higher empirical reward than what can be achieved by this policy. As illustrated in Figure~\ref{fig2}, AdaUCB collects an almost equal reward as the Optimal policy. The cumulative reward achieved by AdaUCB is on the edge of optimality. After a more in-depth look at the results, we noticed that our tests exhibit arms with very low CTRs. Therefore, data logs contain very few clicks compared to a large number of displays. Whether the best arm is selected or not, the clicks are very rare events. Thus, even with perfect CTRs estimation, there a slim probability to generate a reward equal to one in any stage of the experiment. That remains true even without considering the high variance introduced by low CTRs factor. As a result, the bandit algorithms end up with a comparable cumulative reward. Opportunistic bandits are just as statistically valid as other algorithms, and in many circumstances, they can produce answers far more quickly. There are big opportunities offered by an opportunistic bandit algorithm because the latter makes sure that once we learn anything, it is not too late to exploit the best option and avoid the cost resulting from sending influential users to a losing arm.

Finally, we considered another scenario, where the \emph{load} is defined differently from the previous one. We define the \emph{load} as the traffic of customers within a time interval\,\footnote{We set this time interval to 15 minutes.}. Again, this \emph{load} definition is aligned with the main concept of opportunistic bandits: When the \emph{load} is high, the traffic of customers is at peak due to a special promotion, a holiday season or simply a popular time of the day. Thus, we should better avoid any exploration by selecting the arm that we believe is the best. It is a reasonable choice given that retailers strive to offer the best of their services at periods where they are experiencing more crowds than usual. A ``bad'' action in this situation will expose a user to random site experience and can cause him never to come back again, and that seems like a high cost to the business. However, in a period where few users are using the retailer mobile application or website, experimenting with an under-explored arm is a smarter decision since the resulting regret of trying a suboptimal arm will be compensated by the low~\emph{load}. Seasonality is one of the main factors affecting sales and customer demand in the retail industry. In particular, the authors in~\cite{Newing2013} studied the contribution of visitor demand to the seasonal sales variations experienced at grocery retailers, and they demonstrated the significant degree of seasonality experienced around retail stores in terms of their revenue generated from out-of-catchment visitors. For opportunistic bandits, this alternative definition of the \emph{load} naturally captures the potential business opportunities offered by seasonal variations.

\begin{table}[h]
\setlength{\tabcolsep}{5pt}
\centering
\begin{tabular}{@{} l ccc @{}} % @{} serves to suppress white space at ends of table
\toprule
Algorithm   &  \multicolumn{3}{c @{}}{Datasets}\\ 
\cmidrule(l){2-4}
    & {First Test} & {Second Test} & {Third Test}  \\
\midrule
A/B Testing & 3358751 & 5268123 & 916388 \\
DEG     & 226124 & 108641 & 29689  \\
TS      &  361076 & 127366 & 52758 \\
UCB     &  361055 & 145263 & 57021 \\
AdaUCB  &  205715 & 102102 & 25456  \\
\bottomrule
\end{tabular}
\caption{Accumulated Regret at the end of the experiments (Second Scenario).}
\label{tab:second}
\end{table}

\begin{figure}[h] 
	\centering
	\includegraphics[width=\columnwidth]{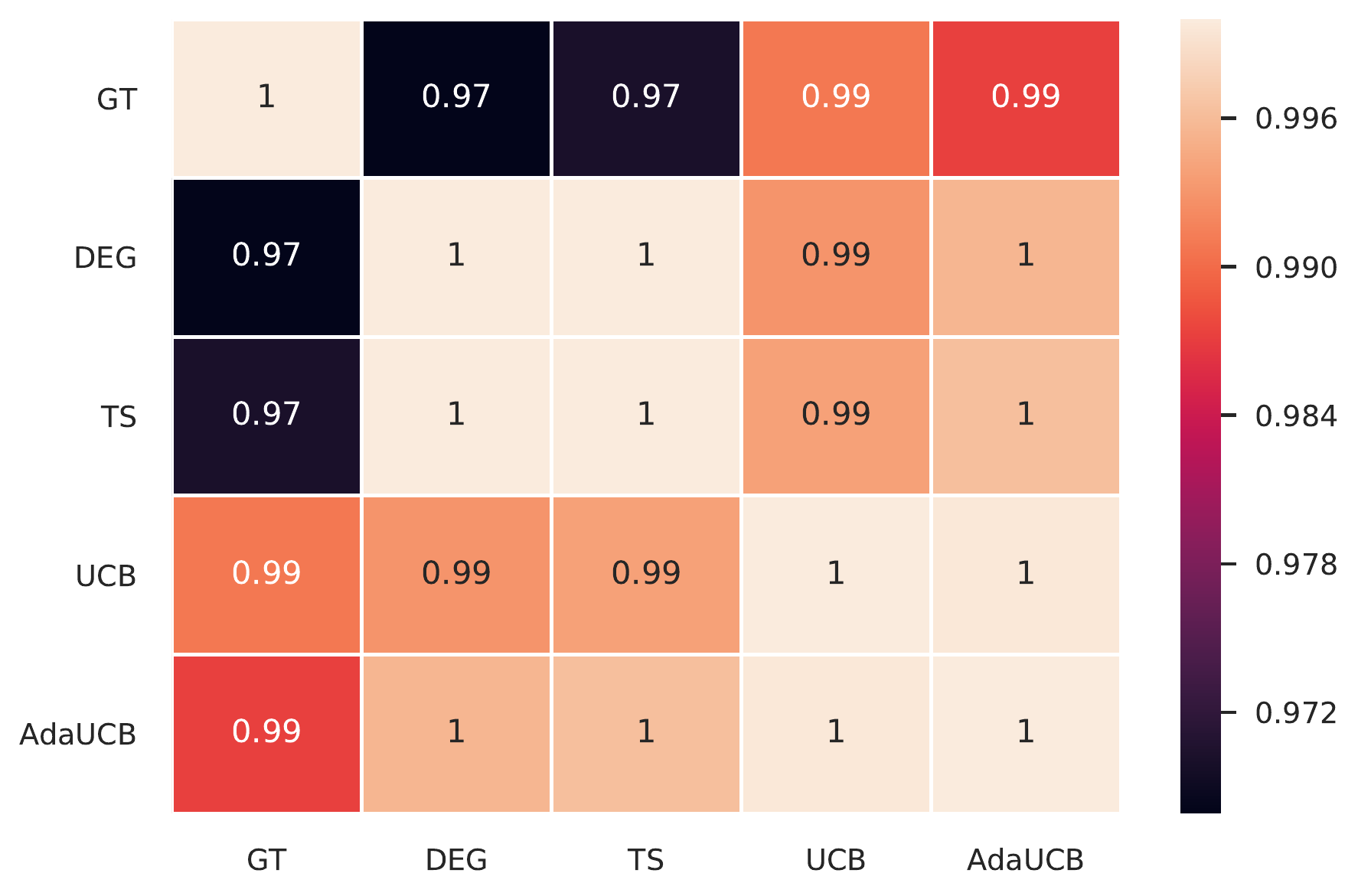}
	\caption{Correlation Matrix for the arms' CTRs values when compared to the Ground Truth (GT) (Second Scenario)}
	\label{fig5}
\end{figure}

The cumulative regret obtained by our evaluated algorithms in Figure~\ref{fig3} and the correlation matrix of CTRs in Figure~\ref{fig5} confirms our previous findings. AdaUCB achieves respectively 9.02\%, 6.01\%, and 14.25\% lower regret than DEG for the three tests. Besides, the regret reduction in compression to UCB is respectively 43.02\%, 29.71\%, and 55.35\% for the three tests. UCB performance is clearly the worst among the other methods because it continues to explore at the expense of exploiting its knowledge of the best arm. DEG appears to be the second-to-best performing by far as expected. DEG policy explores uniformly over all arms. Thus, it will be hurt if there are large number of non-optimal arms. We also notice that its performance degrades rapidly if it is not well tuned. The innovation behind AdaUCB is to ensure that we take into consideration the users' traffic before trying an action, which also warrants that, in the long term, we periodically take a break from exploiting costly arms. The key to opportunistic bandits is its \emph{load}-dependent ``curiosity bonus''. When selecting an arm, AdaUCB calculates the expected reward of each arm and then adds a bonus, which is computed as the inverse proportion to the \emph{load}-dependent confidence of that reward. It is optimistic about uncertainty to the extent of its \emph{load} term. That causes the results of the algorithm to swing wildly from trial to trial based on \emph{load}, especially at the early phases, because each new trial provides more information to the chosen arm so the other arms will mostly be favored more in the subsequent rounds. 

\section{Discussion and Related Work} \label{sec:discussion}
Opportunistic bandits try to account for the data acquisition cost. They do it by balancing the exploration-exploitation trade-off using the current best guess about the performance of each action and a \emph{load}-dependent value of collecting more data to improve upon these guesses. Collecting data will have a real cost, in terms of opportunities lost and their incurred costs as well. Opportunistic bandits were able to achieve superior performance compared to their competitors by pushing explorations to the trials where a qualitative quantity known as \emph{load}, which measures the experiment cost, is low. Technically, these are problems that try to minimize what is known as regret. Regret is the difference between the actual payoff (reward $\times$ \emph{load/cost}) and the payoff you would have collected if you played the best action at every opportunity. 

Generally speaking, bandits experiments terminate earlier than A/B tests because they require a smaller sample. Specifically, they are more conducive to short tests. Indeed, they adjust in real-time by driving more traffic, more quickly, to the better variation. For example, if a retailer is running tests on his website for Black Friday, an A/B test is not practical, given that we might wait until the end of the day to obtain results with high confidence. Because bandits automatically shift traffic to higher performing variations, they represent a low-risk solution for continuous optimization. Finally, they can be used to optimize problems across multiple touchpoints. The communication between bandits ensures that they are working together to optimize the global problem and maximize results. Therefore, bandits might be the best bet when it comes to the multi-version test of web pages describing a complex process or containing revenue-critical elements.

MAB based experimentation has also come under criticism for over-optimizing higher-performing arms. Since bandits algorithms send more traffic to better performing variations, it is likely to reinforce small differences in low-traffic tests and lead to skewed results. This bias is unlikely to generate misleading test results, but it may give a blurry image on how bad is the performance of the worst arms. Opportunistic bandits manage this bias by orchestrating the process of optimizing the best arms or the poorly performing arms depending on \emph{load} conditions. Broadly speaking, the cited problem does not change the fact that MAB tests excel for very short or very long testing cycles, as previously explained. Meanwhile, A/B tests excel for mid-length test cycles when only the variables being tested influence the result given that the other variables are reasonably well understood. Unlike the approach offered by bandit testing, A/B tests force us to wait for a final result until we have a ``significant" data to act on. Therefore, in this case,  it turns out that there is no definite winner, and A/B tests can be better suited to mid-length tests.

In this paper, we experimented with with data that have a relatively small number of arms. One would wonder if AdaUCB continues to be meaningful if we increase the numbers of arms. We believe that we can achieve a larger performance gap with an increased number of arms.  AdaUCB stimulates low-cost discovery.  Introducing more arms makes exploration more costly and lengthy. Given that AdaUCB has the advantage of lower exploration costs compared to its competitors, our strategy is likely to achieve more significant gains.

Compared to traditional bandits algorithms, AdaUCB offers a level of adaptability that none of them can match. It uses \emph{load} factor to weight the competing selections between what currently seems to work best and the potential benefits of collecting more information to improve future results. That creates a built-in way to deal with issues like seasonality that confound other methods. Besides, AdaUCB stimulates low-cost discovery while traditional approaches prefer to reinforce success rather than discover and clarify failure.

\section{Conclusion} \label{sec:conclusion}
Opportunistic bandits can be substantially more efficient optimization methods than traditional multi-armed bandits and other traditional statistical experiments such as A/B testing. Introducing the \emph{load} heuristic for guiding the exploration-exploitation process in multi-armed bandit experiments is simple enough to allow flexible regret optimization. In this paper, we demonstrated that AdaUCB, an opportunistic bandit algorithm, can handle most kinds of issues that arise in real applications of user interface features testing. AdaUCB causes traffic to be dynamically allocated not only based on the superiority of the arms but also based on customer features. One of the significant benefits of AdaUCB is that it mitigates regret, which is basically the lost conversion resulting from exploring a potentially worse variation in a test. By explicitly optimizing the monetary cost of experimentation, opportunistic bandits match the economics of the retail industry much more closely than traditional experiments and should be viewed as the preferred experimental framework. 

\bibliographystyle{siam}
\bibliography{ltexpprt}

\end{document}